\documentclass[conference]{IEEEtran}
\IEEEoverridecommandlockouts
\usepackage{cite}
\usepackage{amsmath,amssymb,amsfonts}
\usepackage{algorithmic}
\usepackage{graphicx}
\usepackage{textcomp}
\usepackage{xcolor}

\usepackage{epstopdf}
\usepackage{multirow}
\usepackage{booktabs}
\usepackage{cuted}
\usepackage{amsmath,lipsum}
\usepackage{xcolor}
\definecolor{mygreen}{HTML}{008000}
\usepackage{siunitx}
\usepackage{amssymb}
\usepackage{array}
\newcolumntype{H}{>{\setbox0=\hbox\bgroup}c<{\egroup}@{}}
\usepackage{colortbl}

\def\BibTeX{{\rm B\kern-.05em{\sc i\kern-.025em b}\kern-.08em
    T\kern-.1667em\lower.7ex\hbox{E}\kern-.125emX}}

\makeatletter
\def\endthebibliography{%
    \def\@noitemerr{\@latex@warning{Empty `thebibliography' environment}}%
    \endlist
}
\makeatother

\usepackage[numbers]{natbib}

\begin{document}

\title{BrainChat: Decoding Semantic Information from fMRI using Vision-language Pretrained Models}

\author{
\IEEEauthorblockN{1\textsuperscript{st} Wanqiu Huang}
\IEEEauthorblockA{\textit{The College of Computer Science and Technology} \\
\textit{Zhejiang University}\\
Hangzhou, China \\
huangwanqiu@zju.edu.cn}\\

\IEEEauthorblockN{3\textsuperscript{rd} Tingyu Xie}
\IEEEauthorblockA{\textit{The College of Computer Science and Technology} \\
\textit{Zhejiang University}\\
Hangzhou, China \\
tingyuxie@zju.edu.cn}

\and

\IEEEauthorblockN{2\textsuperscript{nd} Ke Ma}
\IEEEauthorblockA{\textit{The College of Computer Science and Technology} \\
\textit{Zhejiang University}\\
Hangzhou, China \\
kema@berkeley.edu}\\

\IEEEauthorblockN{4\textsuperscript{th} Hongwei Wang}
\IEEEauthorblockA{\textit{ZJUI} \\
\textit{Zhejiang University}\\
Hangzhou, China \\
hongweiwang@intl.zju.edu.cn}
}

\maketitle

\begin{abstract}
Semantic information is vital for human interaction, and decoding it from brain activity enables non-invasive clinical augmentative and alternative communication.
While there has been significant progress in reconstructing visual images, few studies have focused on the language aspect.
To address this gap, leveraging the powerful capabilities of the decoder-based vision-language pretrained model CoCa, this paper proposes BrainChat, a simple yet effective generative framework aimed at rapidly accomplishing semantic information decoding tasks from brain activity, including fMRI question answering and fMRI captioning.
BrainChat employs the self-supervised approach of Masked Brain Modeling to encode sparse fMRI data, obtaining a more compact embedding representation in the latent space.
Subsequently, BrainChat bridges the gap between modalities by applying contrastive loss, resulting in aligned representations of fMRI, image, and text embeddings.
Furthermore, the fMRI embeddings are mapped to the generative Brain Decoder via cross-attention layers, where they guide the generation of textual content about fMRI in a regressive manner by minimizing caption loss.
Empirically, BrainChat exceeds the performance of existing state-of-the-art methods in the fMRI captioning task and, for the first time, implements fMRI question answering. 
Additionally, BrainChat is highly flexible and can achieve high performance without image data, making it better suited for real-world scenarios with limited data.

\end{abstract}

\begin{IEEEkeywords}
fMRI Decoding, fMRI Captioning, fMRI Question Answer
\end{IEEEkeywords}

\section{Introduction}
\label{sec:introduction}
\begin{figure}
    \centering
    \includegraphics[width=0.4\textwidth]{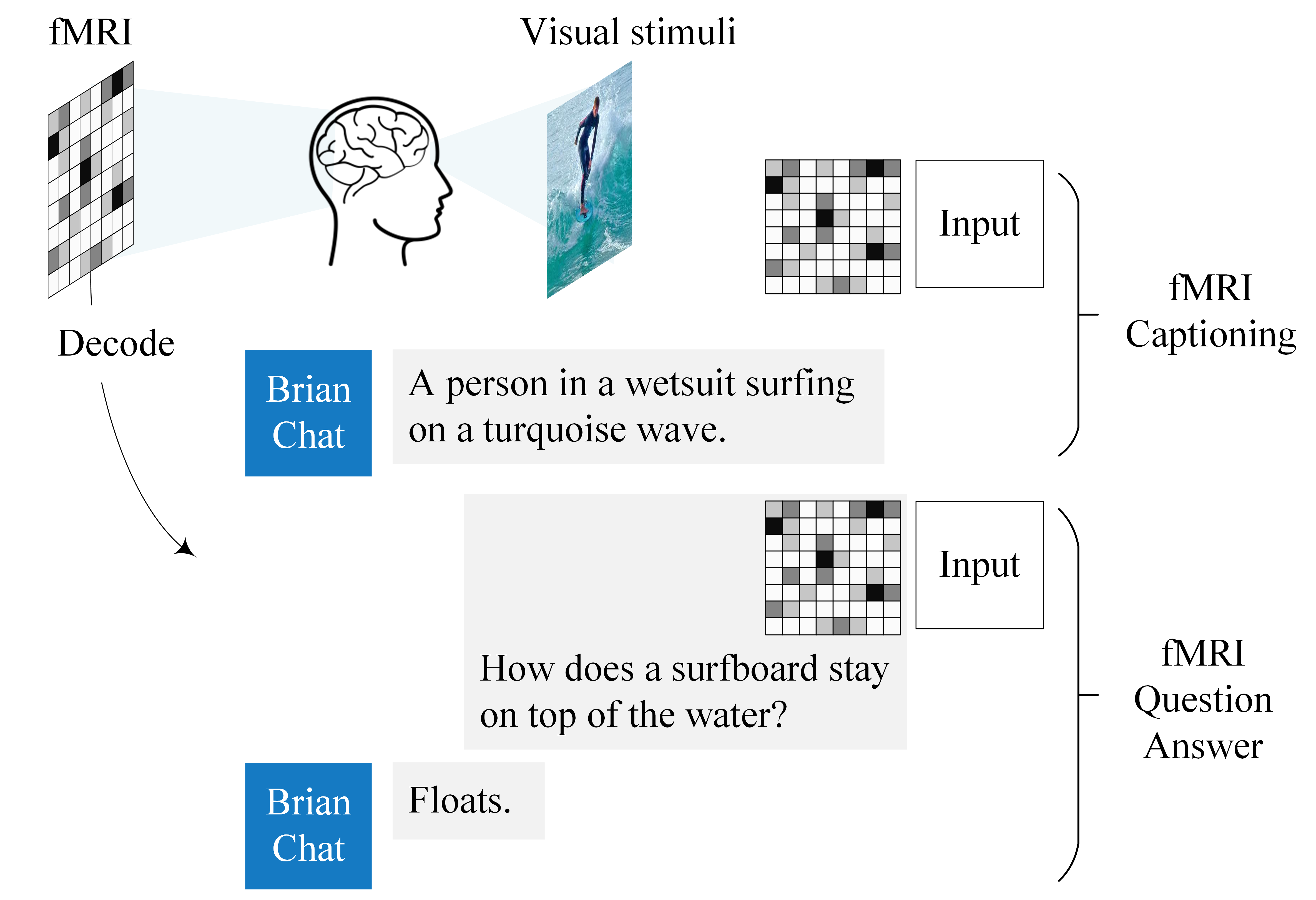}
    \caption{Decoding semantic information from fMRI data. In the fMRI captioning task, a caption is generated based on the fMRI data, describing the semantic information observed during subject scanning. In the fMRI question answering task, corresponding answers are generated based on given questions.}
    \label{fig:intro-tasks}
\end{figure}
\begin{figure*}
        \centering
        \includegraphics[width=0.9\textwidth]{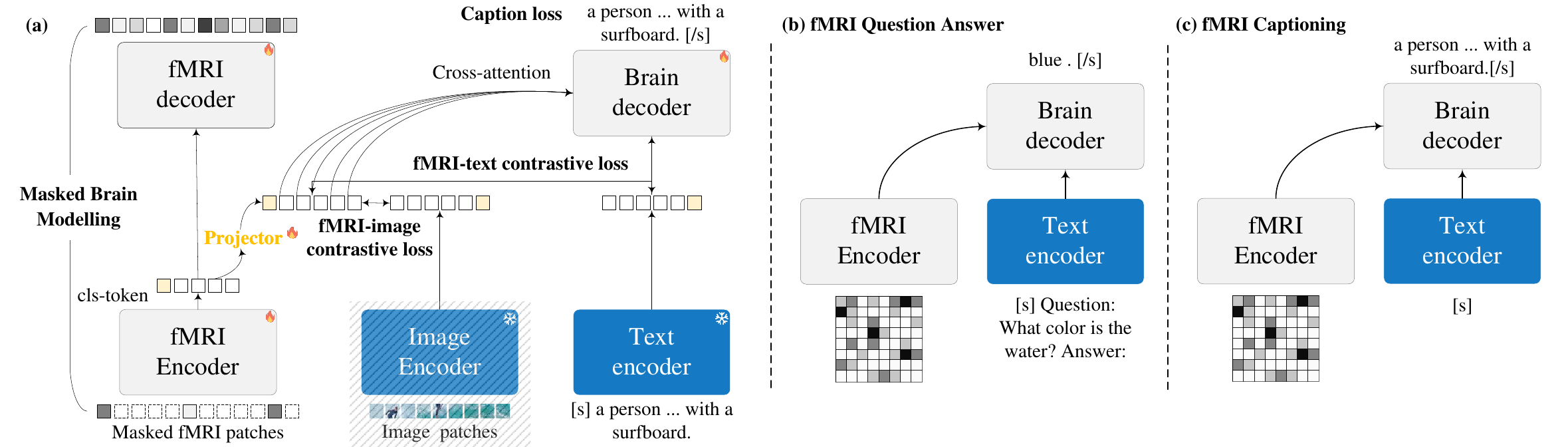}
        \caption{(a) The BrainChat framework consists primarily of encoding and decoding parts. In the encoding part, three encoders are utilized for fMRI, image, and text, each extracting features from its respective modality. The image encoder is exclusively employed during training to enhance the quality of text generation. The decoding part comprises two decoders: the fMRI decoder and the brain decoder, employed to reconstruct masked fMRI data and generate corresponding text based on fMRI features, respectively. During training, we initially reconstruct masked fMRI data using MBM. Subsequently, we train the fMRI encoder and brain decoder using fMRI-image contrastive loss, fMRI-text contrastive loss, and caption loss. Moreover, BrainChat can perform fMRI captioning and fQA tasks without relying on image information, allowing for the removal of the image encoder. (b)(c) In the inference stage, BrainChat is utilized for fQA and fMRI captioning without the need for any visual data.}
        \label{fig:framework}
\end{figure*}

Brain decoding tasks reconstruct the information observed by subjects during scanning from fMRI data.
Many studies have focused on reconstructing visual information from fMRI data, a task known as fMRI-image reconstruction \cite{gazivSelfSupervisedNaturalImage2020,scottiReconstructingMindEye2023,takagiHighresolutionImageReconstruction2023}. However, there has been limited focus on decoding semantic information, another crucial aspect of human communication.

Language serves as a crucial means of communication for humans. Decoding semantic information directly from the brain activity facilitates various clinical applications such as clinical augmentative and alternative communication (AAC) \cite{chaudharySpellingInterfaceUsing2022,maiUniBrainUnifyImage2023} and the functional restoration of aphasia \cite{huangNeuralDecodingAlgorithm2021}.
Similar to image captioning and visual question answering tasks in computer vision, can we use only fMRI data to directly reconstruct the captions corresponding to the images viewed by subjects during scanning (fMRI captioning)? Or to obtain answers to specific questions (fMRI Question Answer)?
Some language decoding works focus on decoding captions or phrases, primarily by extracting fMRI features and then using methods such as LSTM, gated recurrent neural networks, or Transformers to generate caption \cite{huangNeuralDecodingAlgorithm2021,takadaGenerationViewedImage2020,zhangCNNtransformerHybridApproach2022}. 
However, due to framework limitations, existing methods can only generate textual information and cannot facilitate interactive question-answering, which is a more clinically applicable approach.

On the other hand, decoder-based large-scale visual-language models have shown remarkable proficiency in comprehending and generating natural language \cite{liAlignFuseVision2021,baiQwenVLVersatileVisionLanguage2023,liBLIP2BootstrappingLanguageImage2023,wangOFAUnifyingArchitectures2022}.
In this context, Contrastive Captioner (CoCa) \cite{yuCoCaContrastiveCaptioners2022} integrates single-encoder, dual-encoder, and encoder-decoder paradigms in a minimalist design. It pretrains an encoder-decoder foundational model to simultaneously address image-text retrieval and generation tasks.

In this paper, we integrate CoCa with Masked Brain Modeling to establish the generative framework BrainChat. As depicted in Figure~\ref{fig:intro-tasks}, this framework enables fMRI captioning and fMRI question answering (fQA), providing deeper insights into brain activity.
BrainChat adopts a two-stage training approach, comprising a pretraining stage and a brain decoding stage.
During the pretraining stage, BrainChat uses the Masked Brain Modeling method, which involves training an fMRI encoder and an fMRI decoder to reconstruct masked fMRI data, thereby extracting latent representations. Additionally, BrainChat exploits the fixed decoder-based large-scale pretrained model CoCa as the feature extractor for both image and text inputs.
In the brain decoding stage, BrainChat utilizes contrastive loss and caption loss to concurrently optimize the pretrained fMRI encoder and brain decoder, thus achieving text generation based on fMRI.
The main contributions of this paper are as follows:
\begin{enumerate}
\item We have developed a simple yet powerful generative framework named BrainChat for extracting semantic information from fMRI data, facilitating fMRI captioning and fQA tasks.
\item In the fMRI captioning task, our framework surpasses existing diffusion-based state-of-the-art methods. Additionally, we have successfully accomplished the fQA task for the first time, enabling the generation of relevant answers from fMRI data in response to provided questions.
\item In scenarios with restricted data availability, BrainChat demonstrates robust performance when trained exclusively on fMRI-text pairs.
\end{enumerate}

\section{Related Work}
\label{sec:related-work}
\begin{table*}[htbp]
    \caption{Detials of BrainChat. "-" denotes the absence of this parameter.}
    \centering
    \begin{tabular}{@{}ccccccccc@{}}
    \toprule
    \multirow{2}{*}{Model} & \multirow{2}{*}{fMRI encoder} & \multirow{2}{*}{fMRI decoder} & \multicolumn{3}{c}{CoCa-Base}                & \multicolumn{3}{c}{CoCa-Large}               \\ \cmidrule(lr){4-6} \cmidrule(lr){7-9}
                           &                               &                               & Text encoder & Image encoder & Brain decoder & Text encoder & Image encoder & Brain decoder \\ 
    \midrule
    Layers                 & 24                            & 8                             & 12           & 12            & 12            & 12           & 24            & 12            \\
    Hidden                 & 1024                          & 512                           & 512          & 768           & 512           & 768          & 1024          & 768           \\
    Heads                  & 16                            & 16                            & 8            & 12            & 8             & 12           & 16            & 12            \\
    Patch size             & 16                            & -                             & -            & 32            & -             & -            & 14            & -             \\ 
    \bottomrule
    \end{tabular}
    \label{tab:network}
\end{table*}
Existing brain decoding research primarily focuses on reconstructing images, with some also focusing on generating captions from fMRI data. These tasks are commonly referred to as fMRI-image reconstruction and fMRI captioning. However, to the best of our knowledge, there is currently no existing work on fQA, which aims to achieve interactive semantic information decoding.

The fMRI-image reconstruction task is primarily categorized into end-to-end methods and generative-based methods
End-to-end methods \cite{beliyVoxelsPixelsBack2019,gazivSelfSupervisedNaturalImage2020} typically utilize convolutional neural networks (CNNs) to reconstruct images from fMRI data, preserving details but often resulting in blurry images.
On the other hand, generative-based approaches \cite{mozafariReconstructingNaturalScenes2020,scottiReconstructingMindEye2023,takagiHighresolutionImageReconstruction2023,chenSeeingBrainConditional2023,sunContrastAttendDiffuse2023}, leverage models such as Generative Adversarial Networks (GANs) and diffusion models to generate higher-quality images.
In particular, MinD-Vis \cite{chenSeeingBrainConditional2023} introduced Masked Brain Modeling (MBM), which trains fMRI representations using a large pre-training dataset. This approach then reconstructs realistic images from brain recordings with minimal annotations through double-conditioning. Additionally, MindEye \cite{scottiReconstructingMindEye2023} presents an approach for retrieving and reconstructing viewed images from fMRI data, mapping brain activity to high-dimensional latent spaces like CLIP image space for image generation.
Inspired by these works, we incorporate Masked Brain Modeling as our pretraining method and align fMRI features with image and text features simultaneously to decode language from fMRI data.

Research on fMRI captioning tasks is relatively limited, typically involving two main steps: extracting fMRI features and generating captions based on these features. 
\citet{takadaGenerationViewedImage2020} construct a text latent space, where the fMRI data are regressed. Subsequently, they utilize LSTM for decoding to generate captions.
\citet{huangNeuralDecodingAlgorithm2021} propose a language decoding model. It includes an image encoder using a CNN, an fMRI encoder with a bidirectional GRU neural network, and a language decoder based on a GRU neural network for generating phrases or sentences from these features.
\citet{zhangCNNtransformerHybridApproach2022} propose a method where a two-layer 1D CNN converts brain activities into low-dimensional semantic features. Then, the transformer encoder encodes these features into a multi-level abstract representation, and finally, the transformer decoder generates descriptive text about the brain visual stimuli using multi-layer connectivity.
UniBrain \cite{maiUniBrainUnifyImage2023} and MindEye2 \cite{scottiMindEye2SharedSubjectModels2024} map fMRI data to the input space of diffusion models, generating high-quality images and captions. UniBrain employs multi-modal conditions (image and text) to enhance semantic fidelity, while MindEye2 performs multi-subject pretraining, linearly mapping all brain data to a shared-subject latent space.
However, these methods generate text with relatively low quality and fail to consider the question information during text generation. This limitation hinders their ability to effectively accomplish the fQA task.

\section{Method}
\label{method}

As depicted in Figure~\ref{fig:framework}, BrainChat primarily consists of three encoders, one projector, and two decoders. During the pretraining phase, the fMRI encoder and fMRI decoder reconstruct masked fMRI data, extracting latent representations of fMRI. Subsequently, during brain decoding, a projector facilitates aligning fMRI embeddings with CoCa's image and text embeddings, effectively bridging modalities.
Then, the brain decoder generates textual information based on embeddings from three modalities. During this process, we employ contrastive loss and caption loss to train both the fMRI encoder and the brain decoder, enabling text decoding from fMRI data. Specifically, we have implemented architectures similar to those in \cite{yuCoCaContrastiveCaptioners2022,chenSeeingBrainConditional2023}. Both the encoder and decoder models in BrainChat utilize the Vision Transformer \cite{dosovitskiy2021an}. The brain decoder incorporates a causal masking transformer decoder. Additionally, the projector maps fMRI data through an Attentional Pooler.

Training is divided into two stages. During the pretraining phase, we utilize the self-supervised MBM method to pretrain an fMRI encoder and an fMRI decoder as described in \cite{chenSeeingBrainConditional2023}. Initially, the voxels of the fMRI are segmented into patches. Despite masking a portion of the fMRI data, the original information can still be restored due to spatial redundancy. Masking involves setting a specific portion of the patched fMRI to zero. The process of recovering the original data from the remaining masked data aims to minimize noise using MSE loss.
During the Brain Decoding phase, we first employ the pretrained fMRI encoder $f_{\boldsymbol{\theta}}$, along with the fixed image encoder $g_{\boldsymbol{\theta}}$ and the fixed text encoder $q_{\boldsymbol{\theta}}$ from CoCa, to extract embedding features from three data modalities. Subsequently, a projector $p_{\boldsymbol{\theta}}$ is used to map fMRI embeddings to a space with dimensions equivalent to those of image and text embeddings. Following this, we freeze the image encoder and text encoder, and jointly optimize the fMRI encoder and brain decoder using fMRI-image contrastive loss function $L_{fi}$ and fMRI-text contrastive loss function $L_{ft}$:

\begin{strip}
\begin{equation}
    L_{fi} = - (\underbrace{\frac{1}{N}(\sum_i^N \log \frac{\exp (p_{\boldsymbol{\theta}}(f_{\boldsymbol{\theta}}(b_i))^T g_{\boldsymbol{\theta}}(v_i) / \sigma)}{\sum_{j=1}^N \exp (p_{\boldsymbol{\theta}}(f_{\boldsymbol{\theta}}(b_i))^T g_{\boldsymbol{\theta}}(v_j) / \sigma)}}_{\text{fMRI-to-image}}) + \underbrace{\frac{1}{N}(\sum_i^N \log \frac{\exp (g_{\boldsymbol{\theta}}(v_i)^T p_{\boldsymbol{\theta}}(f_{\boldsymbol{\theta}}(b_i)) / \sigma)}{\sum_{j=1}^N \exp (g_{\boldsymbol{\theta}}(v_i)^T f_{\boldsymbol{\theta}}(b_j) / \sigma)})}_{\text{image-to-fMRI}})
\end{equation}
\begin{equation}
    L_{ft} = - (\underbrace{\frac{1}{N}(\sum_i^N \log \frac{\exp (p_{\boldsymbol{\theta}}(f_{\boldsymbol{\theta}}(b_i))^T q_{\boldsymbol{\theta}}(t_i) / \sigma)}{\sum_{j=1}^N \exp (p_{\boldsymbol{\theta}}(f_{\boldsymbol{\theta}}(b_i))^T q_{\boldsymbol{\theta}}(t_j) / \sigma)}}_{\text{fMRI-to-text}}) + \underbrace{\frac{1}{N}(\sum_i^N \log \frac{\exp (q_{\boldsymbol{\theta}}(t_i)^T p_{\boldsymbol{\theta}}(f_{\boldsymbol{\theta}}(b_i)) / \sigma)}{\sum_{j=1}^N \exp (q_{\boldsymbol{\theta}}(t_i)^T f_{\boldsymbol{\theta}}(b_j) / \sigma)})}_{\text{text-to-fMRI}})
\end{equation}
\end{strip}
where $(b_i, v_i, t_i)$ represents paired fMRI-images-text data, with $p_{\boldsymbol{\theta}}(f_{\boldsymbol{\theta}}(b_i))$ and $g_{\boldsymbol{\theta}}(v_j)$ denoting the embeddings of the fMRI data in the $i$-th pair and the image data in the $j$-th pair, respectively. Likewise, $p_{\boldsymbol{\theta}}(t_i)$ denotes the embedding associated with the text data in the $i$-th pair. Here, $N$ signifies the batch size, and $\sigma$ serves as the temperature parameter used for scaling the logits.
We utilize $L_{ft}$ and $L_{fi}$ to concurrently align fMRI data with both images and text, enhancing model robustness and refining the quality of the resulting text generated by the brain decoder. However, for data pairs containing solely fMRI-text $(b_i, t_i)$, aligning fMRI with text data using only $L_{ft}$ yields effective results, as demonstrated in the experimental section.

Moreover, the brain decoder utilizes cross-attention, leveraging fMRI embeddings as conditions and text embeddings as inputs, for generating textual information. The brain decoder learns to maximize the conditional likelihood of the paired text $t$ under the forward autoregressive factorization:

\begin{equation}
    L_{cap} = -\sum_{k=1}^{T}\log P_{\theta}(t_k|t<k,b)
\end{equation}
The caption loss calculates the text $t_k$ generated by the brain decoder at time step $k$, given a caption text across $k$ time steps and fMRI data $b$. Where $ P_{\theta}(t_k|t<k,b) $ represents the probability of generating text $ t_k $ conditioned on previous time steps $ t < k $ and the fMRI data $b$. Therefore, the aim of the caption loss is to minimize the negative log-likelihood of each time step in the generated sequence, ensuring that the generated text closely matches the given caption text.
Ultimately, the total loss function for BrainChat is as follows:
\begin{equation}
    L_{BrainChat} = \lambda_{fi} L_{fi} + \lambda_{ft} L_{ft} + \lambda_{Cap} L_{cap}
\end{equation}

where $\lambda_{fi}$, $\lambda_{ft}$, and $\lambda_{Cap}$ represent the weighted loss hyperparameters. Throughout the training phase, the encoders of the three modalities in BrainChat produce unimodal embeddings, while the brain decoder generates multimodal text representations. 

For fMRI captioning task during training, the brain decoder takes the preceding $k$ text words from the real caption as input, while utilizing the $L_{BrainChat}$ loss to update both the fMRI encoder and the brain decoder simultaneously. In the inference phase, captions are generated through an autoregressive approach, where the output from the brain decoder in the previous time step feeds into the text encoder for the next step, resulting in the predicted caption.

For the fQA task, we fine-tune the fMRI encoder and brain decoder, previously trained on the fMRI captioning task, on the corresponding question-answer dataset using the $L_{cap}$ loss. During inference, similar to the fMRI captioning task, answers are generated using an autoregressive approach, with the question serving as input information for the text encoder through a prompt like "Question: What color is the water? Answer:".

Furthermore, it is worth noting the inherent limitations of fMRI data, which often make it challenging to find corresponding fMRI-image-text data pairs. In such cases, BrainChat can decode semantic information without image data. Within the architecture shown in Figure \ref{fig:framework}, we remove the image encoder and concentrate solely on aligning fMRI embeddings with text embeddings, using only the $L_{ft}$ and $L_{cap}$ loss functions. Despite this simplification, we still achieve strong performance in fMRI captioning and fQA tasks, as detailed in Section~\ref{sec:only_ft}.
\begin{table}[]
    \centering
    \caption{Comparison results of BrainChat with other methods. Here, "M", "R-1", and "R-L" represent METEOR, ROUGE-1, and ROUGE-L, respectively. "-" indicates missing values.}
    \resizebox{0.5\textwidth}{!}{
    \begin{tabular}{cccccc}
        \toprule
        \multirow{2}{*}{Method} & \multirow{2}{*}{Dataset} & \multicolumn{3}{c}{Low-Level} & High-Level \\ 
        \cmidrule(lr){3-5} \cmidrule(lr){6-6}
        & & M$\uparrow$ & R-1$\uparrow$ & R-L$\uparrow$ & CLIP$\uparrow$ \\ 
        \midrule
        \cite{takadaGenerationViewedImage2020} & GOD\cite{horikawaGenericDecodingSeen2017} & - & - & - & 46.4      \\
        \cite{huangNeuralDecodingAlgorithm2021} & \cite{huangNeuralDecodingAlgorithm2021} & - & - & 0.197 & - \\
        \cite{zhangCNNtransformerHybridApproach2022} & \cite{huangNeuralDecodingAlgorithm2021} & - & - & 0.201 & - \\
        UniBrain\cite{maiUniBrainUnifyImage2023} & NSD & 0.169 & 0.245 & 0.222 & 85.3 \\
        MindEye2\cite{scottiMindEye2SharedSubjectModels2024} & NSD & \textbf{0.248} & \textbf{0.326} & 0.353 & 63.8 \\ 
        \midrule
        BrainChat & NSD & 0.155 & 0.3 & \textbf{0.476} & \textbf{91.363} \\ 
        \bottomrule
    \end{tabular}
    }
    \label{tab:comparison_results}
\end{table}

\begin{figure*}[htbp]
    \centering
    \includegraphics[width=0.95\textwidth]{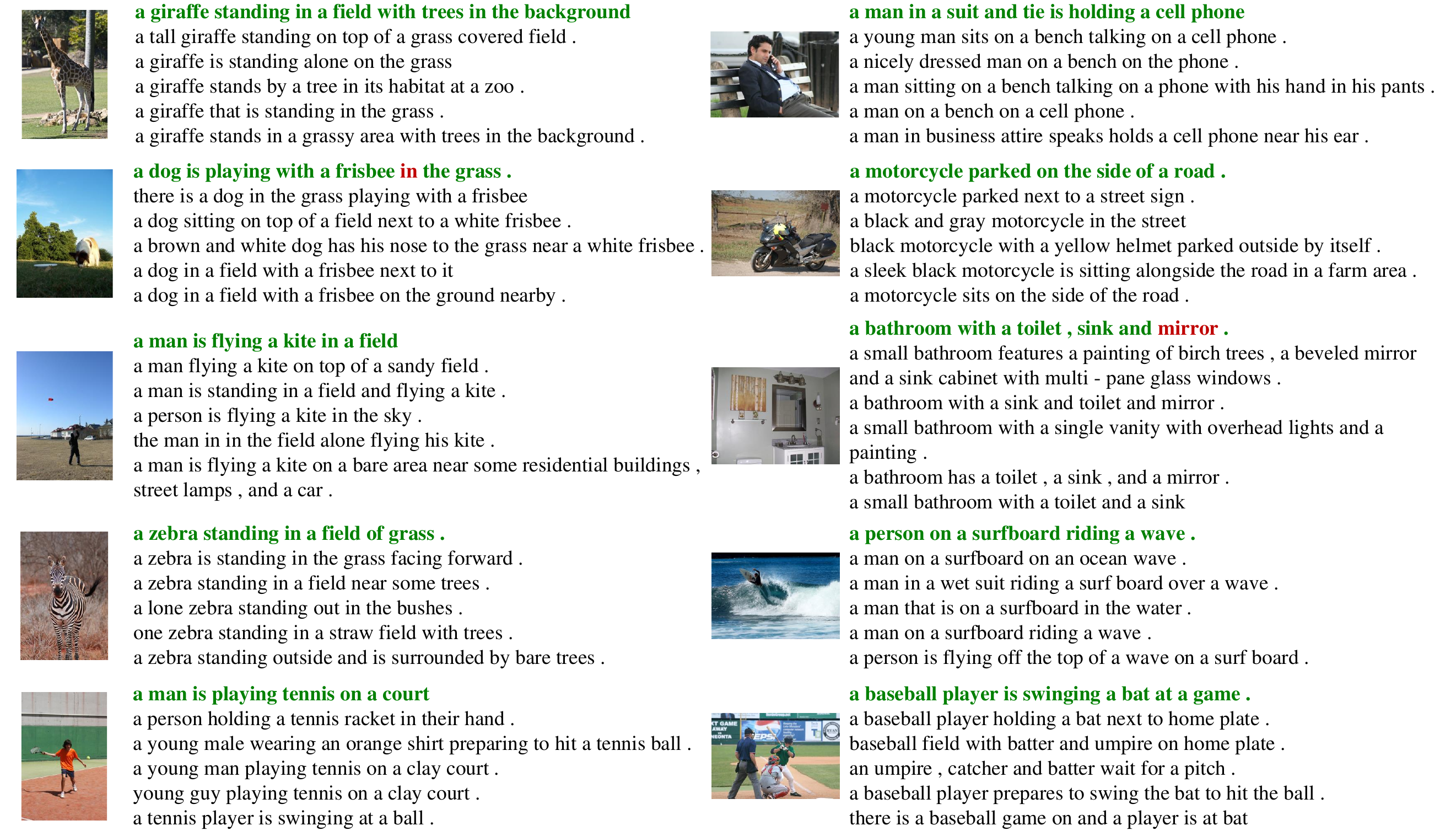}
    \caption{Samples of captions generated by BrainChat. On the left are the visual stimulus images, and on the right are the corresponding captions. \textcolor{mygreen}{Captions generated by BrainChat} are shown in green, while the five ground truth captions are shown in black. It is evident that BrainChat can generate coherent and human-readable text solely using fMRI data. However, there are still some \textcolor{red}{grammar errors} (highlighted in red). Additionally, some of the generated captions are missing periods at the end.}
    \label{fig:captioning}
\end{figure*}

\section{Experiment}

\subsection{Dataset}
The dataset utilized in our paper is the Natural Scenes Dataset (NSD) \cite{allenMassive7TFMRI2022}. This dataset consists of paired fMRI-image data, collected using  a 7T magnetom scanner with a resolution of 1.8 mm. It involves 8 participants who viewed distinct color natural images sourced from the COCO dataset \cite{lin2014microsoft}. Consistent with prior studies \cite{scottiReconstructingMindEye2023, takagiHighresolutionImageReconstruction2023}, we followed the same standardized train/test splits. The test dataset consists of 982 samples, while the training dataset comprises 24980 samples. And we utilized data from subject 1, which included 15,724 voxels.
For the fMRI captioning task, we combined the NSD dataset with the captioning data from the COCO dataset to construct fMRI-image-text paired data. 
During the training process, we have the flexibility to use either all three types of information as input, or only fMRI-text data.
For the fQA task, we integrated the NSD dataset with questions, answers, and images sourced from the VQA dataset \cite{balanced_vqa_v2}, where the images in the VQA dataset are also obtained from COCO.

\subsection{Implementation}
In BrainChat, since CoCa is not open-sourced, we utilized a pre-trained CoCa model provided by Openclip \cite{ilharco_gabriel_2021_5143773}. All encoders and decoders in our architecture adopt the ViT architecture. Additional specifications are outlined in Table~\ref{tab:network}. We employed a large mask ratio of 0.75 to mask the fMRI data in the NSD dataset, following the method proposed in \cite{chenSeeingBrainConditional2023}. 
Additionally, during the pretraining phase, the learning rate was set to 5e-5, and the weight decay was set to 0.05, determined by random search. The AdamW optimizer was utilized with $\beta_1=0.9$ and $\beta_2=0.95$, along with the NativeScaler for scaling gradients.
During the brain decoding phase, the training parameters were determined through random search as follows: a learning rate of 1e-4, weight decay of 0.1, caption loss weight of 20.0, and contrastive loss weight of 1.0. We also incorporated a trainable temperature parameter $\tau$ for the contrastive loss, following in Openclip. Additionally, we explored the performance of BrainChat under different pretraining and fine-tuning settings in Section~\ref{sec:quantitative_captioning}.

\subsection{Results of fMRI Captioning}
\label{sec:qualitative_captioning}
Figure~\ref{fig:captioning} shows the captions generated by BrainChat. These samples were produced using a BrainChat model pretrained on the HCP and NSD datasets. During the brain decoding fine-tuning phase, the fMRI encoder was partially frozen, with the last 10 blocks being trained. The input data was fMRI, and the model autoregressively generated the corresponding captions. As shown in Figure~\ref{fig:captioning}, BrainChat can generate captions that match the semantic information of the visual stimuli corresponding to the fMRI. Except for a few minor errors, all sentences are generally grammatically correct and can be read fluently by humans.

\subsection{fMRI Captioning Evaluation}
\label{sec:quantitative_captioning}
To conduct quantitative comparisons, we employed the same low-level and high-level evaluation metrics as those used in \cite{maiUniBrainUnifyImage2023,scottiMindEye2SharedSubjectModels2024}, including METEOR, ROUGE-1, ROUGE-L, and the 2-way CLIP metric. The 2-way CLIP metric assesses whether the original fMRI embedding is more similar to its corresponding text embedding or to a randomly selected one.
We evaluated the quality of captions generated by BrainChat using the same pretraining and fine-tuning settings as described in Section~\ref{sec:qualitative_captioning}. The comparison with four other methods is presented in Table~\ref{tab:comparison_results}.
Due to the limited number of existing fMRI captioning works, we conducted a preliminary comparison with the first three methods \cite{takadaGenerationViewedImage2020,huangNeuralDecodingAlgorithm2021,zhangCNNtransformerHybridApproach2022}, as done in \cite{maiUniBrainUnifyImage2023}.
As shown in Table~\ref{tab:eval_results}, BrainChat performs well on both low-level and high-level evaluation metrics. Particularly, BrainChat outperformed UniBrain in terms of ROUGE-1, ROUGE-L, and CLIP across the four evaluation metrics. Under high-level metrics, it demonstrated a 27.6\% improvement over MindEye2.

We also assessed BrainChat's caption generation using different CoCa model variants and pretraining and fine-tuning setups. We used two CoCa scales: CoCa-Base (CoCa-B) and CoCa-Large (CoCa-L). We explored scenarios where pretraining occurred on both the HCP and NSD datasets (referred to as "all"), solely on the HCP dataset (referred to as "w/o NSD"), and without pretraining (referred to as "none"). During brain decoding training, we evaluated the fMRI encoder's performance with all parameters frozen (referred to as "frozen whole") and partially frozen (referred to as "frozen partly").
Moreover, we included additional common text evaluation metrics such as BLEU@1, BLEU@2, BLEU@3, BLEU@4, and CIDEr.
The experimental results are shown in Table~\ref{tab:eval_results}, with the variance from five repeated evaluations available in the appendix.

From the table, it can be seen that ID 7, which uses CoCa-Large, is pretrained on both the NSD and HCP datasets, and finetunes part of the fMRI encoder, achieved the best results in 4 out of the 9 metrics.  
Additionally, the network performance under different settings can be summarized as follows: (1) For the high-level CLIP metric, the ID 5-10 experiments, which utilized CoCa-Large, showed superior performance. (2) When other settings were consistent, pretraining on both the HCP and NSD datasets resulted in better outcomes, as observed in the ID 1, 3, and 5 experiments, among others. (3) Freezing parts of the fMRI encoder network during fine-tuning led to enhanced results, as evidenced by the ID 1 and 2 experiments, among others.

\begin{table*}[htbp]
    \centering
    \caption{Quantitative evaluation results of fMRI captioning. ID 1-10 represent BrainChat's performance under various settings. Experiments with IDs 5 and 10 were not pretrained, and thus no network parameters were frozen during the fine-tuning stage. The experiments with a \colorbox{gray!20}{gray background} indicate our default setting.}
    \begin{tabular}{ccccccccclllllllll}
        \toprule
        \multirow{2}{*}{ID} & \multicolumn{2}{c}{CoCa} & \multicolumn{3}{c}{Pretrain} & \multicolumn{2}{c}{Finetune} & \multirow{2}{*}{B@1$\uparrow$} & \multirow{2}{*}{B@2$\uparrow$} & \multirow{2}{*}{B@3$\uparrow$} & \multirow{2}{*}{B@4$\uparrow$} & \multirow{2}{*}{M$\uparrow$} & \multirow{2}{*}{R-1$\uparrow$} & \multirow{2}{*}{R-L$\uparrow$} & \multirow{2}{*}{CIDEr$\uparrow$} & \multirow{2}{*}{CLIP$\uparrow$} \\ 
        \cmidrule(lr){2-8}
            & B & L & all & w/o NSD & none & none & partly & & & & & & & & & \\
        \midrule
        1 & \checkmark &  & \checkmark &  &  & \checkmark &  & 0.518 & 0.287 & 0.169 & 0.107 & 0.144 & 0.283 & 0.459 & 0.236 & 63.492 \\
        2 & \checkmark &  & \checkmark &  &  &  & \checkmark & 0.530 & 0.307 & 0.186 & 0.120 & \textbf{0.159} & 0.288 & 0.461 & \textbf{0.322} & 86.225 \\
        3 & \checkmark &  &  & \checkmark &  & \checkmark &  & 0.468 & 0.239 & 0.130 & 0.077 & 0.130 & 0.264 & 0.424 & 0.154 & 63.942 \\
        4 & \checkmark &  &  & \checkmark &  &  & \checkmark & 0.518 & 0.298 & 0.180 & 0.118 & 0.152 & 0.288 & 0.454 & 0.299 & 85.896 \\
        5 & \checkmark &  &  &  & \checkmark &  &  & 0.452 & 0.214 & 0.105 & 0.058 & 0.125 & 0.242 & 0.411 & 0.118 & 72.625 \\
        6 &  & \checkmark & \checkmark &  &  & \checkmark &  & 0.509 & 0.272 & 0.153 & 0.094 & 0.145 & 0.275 & 0.445 & 0.236 & 90.163 \\
        \rowcolor{gray!20} 7 &  & \checkmark & \checkmark &  &  &  & \checkmark & \textbf{0.555} & 0.316 & \textbf{0.191} & \textbf{0.124} & 0.155 & \textbf{0.300} & \textbf{0.476} & 0.307 & 91.363 \\
        8 &  & \checkmark &  & \checkmark &  & \checkmark &  & 0.499 & 0.249 & 0.125 & 0.069 & 0.129 & 0.268 & 0.434 & 0.144 & 80.204 \\
        9 &  & \checkmark &  & \checkmark &  &  & \checkmark & 0.550 & \textbf{0.317} & 0.187 & 0.113 & 0.153 & 0.299 & 0.471 & 0.298 & \textbf{92.633} \\
        10 &  & \checkmark &  &  & \checkmark &  &  & 0.483 & 0.258 & 0.145 & 0.090 & 0.135 & 0.266 & 0.438 & 0.175 & 88.342 \\ 
        \midrule
        \multicolumn{8}{c}{UniBrain\cite{maiUniBrainUnifyImage2023}} & - & - & - & - & 0.169 & 0.245 & 0.222 & - & 85.3 \\ 
        \multicolumn{8}{c}{MindEye2\cite{scottiMindEye2SharedSubjectModels2024}} & - & - & - & - & \textbf{0.248} & \textbf{0.326} & 0.353 & - & 63.8 \\ 
        \bottomrule
    \end{tabular}
    \label{tab:eval_results}
\end{table*}

\begin{table}[htbp]
    \centering
    \caption{Accuracy of Brainchat in fQA tasks under different settings.}
    \resizebox{0.5\textwidth}{!}{
    \begin{tabular}{lllllllll}
    \toprule
    \multicolumn{1}{c}{ID} & \multicolumn{2}{c}{CoCa} & \multicolumn{3}{c}{Pretrain} & \multicolumn{2}{c}{Finetune} & \multicolumn{1}{c}{\multirow{2}{*}{Acc}} \\ \cmidrule{2-8}
    \multicolumn{1}{c}{} & \multicolumn{1}{c}{B} & \multicolumn{1}{c}{L} & \multicolumn{1}{c}{all} & \multicolumn{1}{c}{w/o NSD} & \multicolumn{1}{c}{none} & \multicolumn{1}{c}{none} & \multicolumn{1}{c}{partly} & \multicolumn{1}{c}{} \\
    \midrule
    1 & \checkmark &  & \checkmark &  &  & \checkmark &  & 0.405 \\
    2 & \checkmark &  & \checkmark &  &  &  & \checkmark & 0.411 \\
    3 & \checkmark &  &  & \checkmark &  & \checkmark &  & 0.393 \\
    4 & \checkmark &  &  & \checkmark &  &  & \checkmark & 0.413 \\
    5 & \checkmark &  &  &  & \checkmark &  &  & 0.391 \\
    6 &  & \checkmark & \checkmark &  &  & \checkmark &  & 0.398 \\
    7 &  & \checkmark & \checkmark &  &  &  & \checkmark & 0.385 \\
    8 &  & \checkmark &  & \checkmark &  & \checkmark &  & 0.387 \\
    9 &  & \checkmark &  & \checkmark &  &  & \checkmark & 0.386 \\
    10 &  & \checkmark &  &  & \checkmark &  &  & \textbf{0.417} \\ 
    \bottomrule
    \end{tabular}
    }
    \label{tab:fQA_acc}
\end{table}

\begin{figure*}[htbp]
    \centering
    \includegraphics[width=1\textwidth]{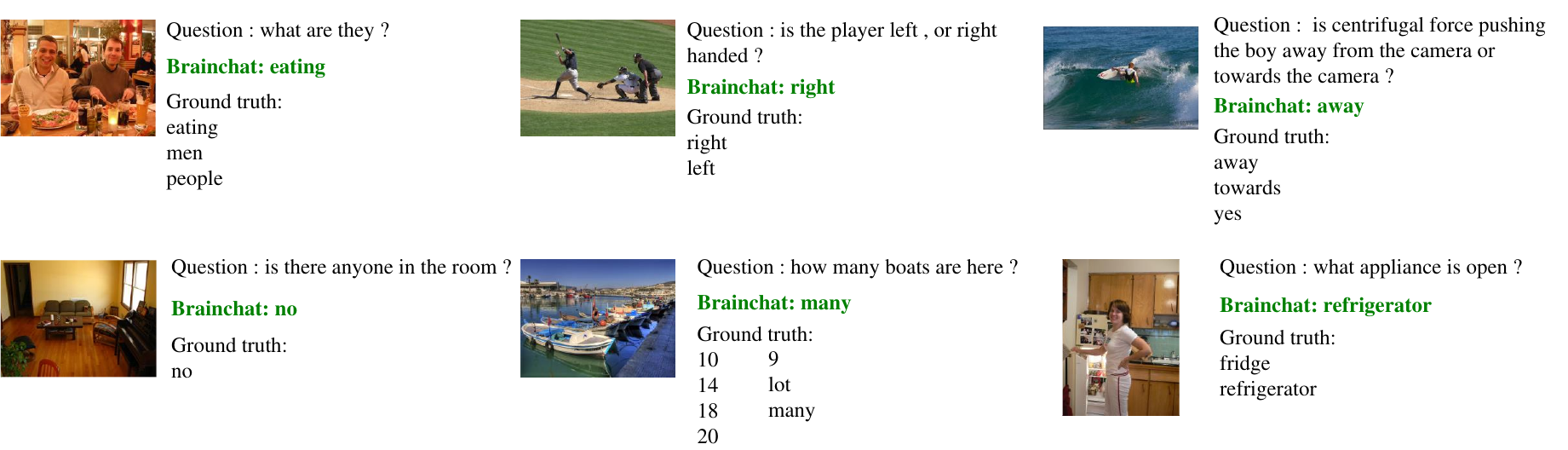}
    \caption{Samples of fQA output by Brainchat. The text in green represents \textcolor{mygreen}{the answers generated by Brainchat} based on the questions. Ground truth refers to all the answers provided by the VQA dataset. }
    \label{fig:fQA}
\end{figure*}

\begin{table*}[htbp]
    \caption{ID 1-10 shows the results of training Brainchat using only fMRI-text data pairs, which are the quantitative evaluation results for Brainchat on the fMRI captioning task. For comparison, the last row of the table lists the results of the ID 7 experiment in Table \ref{tab:eval_results}, which used fMRI-image-text data pairs.}
    \begin{tabular}{lccccccclllllllll}
    \toprule
    \multirow{2}{*}{ID} & \multicolumn{2}{c}{CoCa} & \multicolumn{3}{c}{Pretrain} & \multicolumn{2}{c}{Finetune} & \multirow{2}{*}{B@1$\uparrow$} & \multirow{2}{*}{B@2$\uparrow$} & \multirow{2}{*}{B@3$\uparrow$} & \multirow{2}{*}{B@4$\uparrow$} & \multirow{2}{*}{M$\uparrow$} & \multirow{2}{*}{R-1$\uparrow$} & \multirow{2}{*}{R-L$\uparrow$} & \multirow{2}{*}{CIDEr$\uparrow$} & \multirow{2}{*}{CLIP$\uparrow$} \\ 
    \cmidrule(lr){2-8}
    & B & L & all & w/o NSD & none & none & partly & & & & & & & & & \\
    \midrule
    1 & \checkmark &  & \checkmark &  &  & \checkmark &  & 0.523 & 0.292 & 0.171 & 0.107 & 0.143 & 0.285 & 0.457 & 0.261 & 63.300 \\
    2 & \checkmark &  & \checkmark &  &  &  & \checkmark & \textbf{0.539} & 0.308 & 0.183 & 0.118 & 0.153 & 0.283 & 0.467 & 0.296 & 86.983 \\
    3 & \checkmark &  &  & \checkmark &  & \checkmark &  & 0.466 & 0.231 & 0.123 & 0.073 & 0.128 & 0.256 & 0.425 & 0.159 & 64.654 \\
    4 & \checkmark &  &  & \checkmark &  &  & \checkmark & 0.523 & 0.297 & 0.177 & 0.113 & 0.151 & 0.289 & 0.458 & 0.291 & 88.179 \\
    5 & \checkmark &  &  &  & \checkmark &  &  & 0.508 & 0.243 & 0.125 & 0.077 & 0.127 & 0.274 & 0.445 & 0.090 & 50.413 \\
    6 &  & \checkmark & \checkmark &  &  & \checkmark &  & 0.538 & \textbf{0.310} & 0.184 & 0.118 & 0.148 & 0.290 & 0.467 & 0.274 & 81.383 \\
    \rowcolor{gray!20} 7 &  & \checkmark & \checkmark &  &  &  & \checkmark & 0.528 & 0.301 & \textbf{0.185} & \textbf{0.119} & \textbf{0.157} & \textbf{0.294} & \textbf{0.470} & \textbf{0.314} & \textbf{94.175} \\
    8 &  & \checkmark &  & \checkmark &  & \checkmark &  & 0.473 & 0.238 & 0.126 & 0.072 & 0.124 & 0.259 & 0.423 & 0.152 & 75.121 \\
    9 &  & \checkmark &  & \checkmark &  &  & \checkmark & 0.532 & 0.305 & 0.181 & 0.115 & 0.155 & 0.293 & 0.466 & 0.284 & 92.638 \\
    10 &  & \checkmark &  &  & \checkmark &  &  & 0.469 & 0.229 & 0.102 & 0.055 & 0.117 & 0.257 & 0.422 & 0.081 & 77.033 \\
    \midrule
    \multicolumn{8}{c}{Result of ID 7 experiment in Table~\ref{tab:eval_results}} & \textbf{0.555} & \textbf{0.316} & \textbf{0.191} & \textbf{0.124} & 0.155 & \textbf{0.300} & \textbf{0.476} & 0.307 & 91.363 \\
    \bottomrule
    \end{tabular}
    \label{tab:captioning_ft}
\end{table*}

\subsection{Results of fQA}
To adapt BrainChat for the fQA task, we utilized the BrainChat model trained on the fMRI captioning task as described in Section~\ref{sec:qualitative_captioning}, and further fine-tuned it on the VQA dataset.
For the VQA fine-tuning process, we mainly refer to the settings in \cite{yuCoCaContrastiveCaptioners2022}. We convert the fQA task into a classification task by training a linear classifier on the output layer of the brain decoder to predict the final answer. Specifically, the question is first tokenized, and then the tokens and the fMRI data are respectively input into the text encoder and fMRI encoder to obtain the corresponding embeddings. The text embedding is input into the brain decoder, while the fMRI embedding serves as a condition in the brain decoder's cross-attention module, generating output tokens. Finally, we retain only the token predicted by the brain decoder based on the question tokens, and input it into the linear classifier to obtain the final prediction.
During the fine-tuning process with the VQA dataset, we used fMRI-question-answer pairs as input data, optimizing the linear classifier with Cross Entropy Loss. The optimizer was AdamW with a learning rate of 1e-6 and a weight decay of 0.1. 

As shown in the Figure~\ref{fig:fQA}, there are samples of fQA task from Brainchat. It demonstrates that Brainchat can consider both fMRI and question information to provide the specified answer. However, the red incorrect samples indicate that Brainchat still needs improvement in capturing detailed information from fMRI data.

\subsection{fQA Evaluation}
We assessed BrainChat's performance across various settings in the fQA. Initially, we employed the Brainchat model trained under ten different settings in the fMRI captioning, as detailed in Section~\ref{sec:quantitative_captioning}, and subsequently finetuned it on the VQA dataset.
We utilized accuracy as our evaluation metric, with results presented in the Table~\ref{tab:fQA_acc}. It is shown that the various settings had minimal impact on the final fQA results, with the accuracy clustering between 0.38 and 0.41.
Compared to the fMRI captioning task, this emphasizes BrainChat's ability to effectively handle the fQA task across various settings. This further underscores the potential robustness of the BrainChat model. By extracting effective representations from the initial fMRI captioning task, the model seamlessly transferred these representations to the fQA task.
Otherwise, experiment ID 10, leveraging the CoCa-Large model without any pretraining on datasets, achieved the highest accuracy of 0.417.

\subsection{Adapting BrainChat to Limited fMRI Data}
\label{sec:only_ft}
Due to the scarcity of fMRI data, obtaining corresponding fMRI-image-text data pairs may not be feasible. BrainChat can perform fMRI captioning and fQA using only fMRI-text data pairs. Specifically, during the brain decoding phase, only the fMRI encoder, text encoder, and brain decoder are utilized, with optimization performed using $L_{ft}$ and $L_{cap}$. All other training parameters remain unchanged.
The evaluation metrics for fMRI captioning under different settings are shown in the Table~\ref{tab:captioning_ft}.

The table shows that Brainchat exhibits little variation across ten different settings on eight low-level evaluation metrics. However, the ID 5-10 experiments using the CoCa-Large model perform better on the high-level CLIP metric. Additionally, the ID 7 experiment, which used the CoCa-Large model, was pretrained on the HCP and NSD datasets, and partially froze the fMRI encoder during the brain decoding fine-tuning phase, achieved superior results, excelling in 7 out of 9 metrics.
The results from the ID 7 experiment in Table \ref{tab:eval_results}, trained on all three modalities, enhance fMRI captioning performance across 6 metrics. However, even when utilizing only fMRI-text data, BrainChat maintains strong performance. This suggests that BrainChat's performance remains robust even with limited fMRI data.

\section{Conclusion}
This paper introduces Brainchat, a semantic information decoding framework designed to extract textual information from fMRI data, facilitating fMRI captioning and fQA tasks. For fMRI captioning, we adopt a two-stage training strategy. Brainchat surpasses existing state-of-the-art methods in the fMRI captioning task. Furthermore, we analyze the performance of Brainchat under different pretraining and finetuning settings.
In the fQA task, the Brain Decoder utilizes cross-attention, employing fMRI embeddings as conditions to generate answers, enabling fQA for the first time.
We hope that Brainchat not only pushes the boundaries of brain decoding technology but also opens up possibilities for clinical applications such as augmentative and alternative communication (AAC) and human-computer interaction.

BrainChat demonstrates strong extensibility. In our future work, we aim to integrate diffusion to establish a unified framework for fMRI decoding in clinical applications. Additionally, we will investigate enhanced cross-attention mechanisms to improve comprehension of fMRI embeddings.

\bibliographystyle{IEEEtranN}
\bibliography{bibfile}

\begin{thebibliography}{24}
\providecommand{\natexlab}[1]{#1}
\providecommand{\url}[1]{#1}
\csname url@samestyle\endcsname
\providecommand{\newblock}{\relax}
\providecommand{\bibinfo}[2]{#2}
\providecommand{\BIBentrySTDinterwordspacing}{\spaceskip=0pt\relax}
\providecommand{\BIBentryALTinterwordstretchfactor}{4}
\providecommand{\BIBentryALTinterwordspacing}{\spaceskip=\fontdimen2\font plus
\BIBentryALTinterwordstretchfactor\fontdimen3\font minus \fontdimen4\font\relax}
\providecommand{\BIBforeignlanguage}[2]{{%
\expandafter\ifx\csname l@#1\endcsname\relax
\typeout{** WARNING: IEEEtranN.bst: No hyphenation pattern has been}%
\typeout{** loaded for the language `#1'. Using the pattern for}%
\typeout{** the default language instead.}%
\else
\language=\csname l@#1\endcsname
\fi
#2}}
\providecommand{\BIBdecl}{\relax}
\BIBdecl

\bibitem[Gaziv et~al.(2020)Gaziv, Beliy, Granot, Hoogi, Strappini, Golan, and Irani]{gazivSelfSupervisedNaturalImage2020}
G.~Gaziv, R.~Beliy, N.~Granot, A.~Hoogi, F.~Strappini, T.~Golan, and M.~Irani, ``Self-{{Supervised Natural Image Reconstruction}} and {{Rich Semantic Classification}} from {{Brain Activity}},'' p. 2020.09.06.284794, Sep. 2020.

\bibitem[Scotti et~al.(2023)Scotti, Banerjee, Goode, Shabalin, Nguyen, Cohen, Dempster, Verlinde, Yundler, Weisberg, Norman, and Abraham]{scottiReconstructingMindEye2023}
P.~S. Scotti, A.~Banerjee, J.~Goode, S.~Shabalin, A.~Nguyen, E.~Cohen, A.~J. Dempster, N.~Verlinde, E.~Yundler, D.~Weisberg, K.~A. Norman, and T.~M. Abraham, ``Reconstructing the {{Mind}}'s {{Eye}}: {{fMRI-to-Image}} with {{Contrastive Learning}} and {{Diffusion Priors}},'' May 2023.

\bibitem[Takagi and Nishimoto(2023)]{takagiHighresolutionImageReconstruction2023}
Y.~Takagi and S.~Nishimoto, ``High-resolution image reconstruction with latent diffusion models from human brain activity,'' p. 2022.11.18.517004, Mar. 2023.

\bibitem[Chaudhary et~al.(2022)Chaudhary, Vlachos, Zimmermann, Espinosa, Tonin, {Jaramillo-Gonzalez}, {Khalili-Ardali}, Topka, Lehmberg, Friehs, Woodtli, Donoghue, and Birbaumer]{chaudharySpellingInterfaceUsing2022}
U.~Chaudhary, I.~Vlachos, J.~B. Zimmermann, A.~Espinosa, A.~Tonin, A.~{Jaramillo-Gonzalez}, M.~{Khalili-Ardali}, H.~Topka, J.~Lehmberg, G.~M. Friehs, A.~Woodtli, J.~P. Donoghue, and N.~Birbaumer, ``Spelling interface using intracortical signals in a completely locked-in patient enabled via auditory neurofeedback training,'' \emph{Nature Communications}, vol.~13, no.~1, p. 1236, 2022.

\bibitem[Mai and Zhang(2023)]{maiUniBrainUnifyImage2023}
W.~Mai and Z.~Zhang, ``{{UniBrain}}: {{Unify Image Reconstruction}} and {{Captioning All}} in {{One Diffusion Model}} from {{Human Brain Activity}},'' Aug. 2023.

\bibitem[Huang et~al.(2021)Huang, Yan, Cheng, Wang, Li, Wang, Li, Li, Li, Zuo, and Chen]{huangNeuralDecodingAlgorithm2021}
W.~Huang, H.~Yan, K.~Cheng, C.~Wang, J.~Li, Y.~Wang, C.~Li, C.~Li, Y.~Li, Z.~Zuo, and H.~Chen, ``A neural decoding algorithm that generates language from visual activity evoked by natural images,'' \emph{Neural Networks}, vol. 144, pp. 90--100, Dec. 2021.

\bibitem[Takada et~al.(2020)Takada, Togo, Ogawa, and Haseyama]{takadaGenerationViewedImage2020}
S.~Takada, R.~Togo, T.~Ogawa, and M.~Haseyama, ``Generation of {{Viewed Image Captions From Human Brain Activity Via Unsupervised Text Latent Space}},'' in \emph{2020 {{IEEE International Conference}} on {{Image Processing}} ({{ICIP}})}, Oct. 2020, pp. 2521--2525.

\bibitem[Zhang et~al.(2022)Zhang, Li, Liu, Min, Wang, Li, Wang, Yan, Zuo, Huang, and Chen]{zhangCNNtransformerHybridApproach2022}
J.~Zhang, C.~Li, G.~Liu, M.~Min, C.~Wang, J.~Li, Y.~Wang, H.~Yan, Z.~Zuo, W.~Huang, and H.~Chen, ``A {{CNN-transformer}} hybrid approach for decoding visual neural activity into text,'' \emph{Computer Methods and Programs in Biomedicine}, vol. 214, p. 106586, Feb. 2022.

\bibitem[Li et~al.(2021)Li, Selvaraju, Gotmare, Joty, Xiong, and Hoi]{liAlignFuseVision2021}
J.~Li, R.~R. Selvaraju, A.~D. Gotmare, S.~Joty, C.~Xiong, and S.~Hoi, ``Align before {{Fuse}}: {{Vision}} and {{Language Representation Learning}} with {{Momentum Distillation}},'' Oct. 2021.

\bibitem[Bai et~al.(2023)Bai, Bai, Yang, Wang, Tan, Wang, Lin, Zhou, and Zhou]{baiQwenVLVersatileVisionLanguage2023}
J.~Bai, S.~Bai, S.~Yang, S.~Wang, S.~Tan, P.~Wang, J.~Lin, C.~Zhou, and J.~Zhou, ``Qwen-{{VL}}: {{A Versatile Vision-Language Model}} for {{Understanding}}, {{Localization}}, {{Text Reading}}, and {{Beyond}},'' Oct. 2023.

\bibitem[Li et~al.(2023)Li, Li, Savarese, and Hoi]{liBLIP2BootstrappingLanguageImage2023}
J.~Li, D.~Li, S.~Savarese, and S.~Hoi, ``{{BLIP-2}}: {{Bootstrapping Language-Image Pre-training}} with {{Frozen Image Encoders}} and {{Large Language Models}},'' Jun. 2023.

\bibitem[Wang et~al.(2022)Wang, Yang, Men, Lin, Bai, Li, Ma, Zhou, Zhou, and Yang]{wangOFAUnifyingArchitectures2022}
P.~Wang, A.~Yang, R.~Men, J.~Lin, S.~Bai, Z.~Li, J.~Ma, C.~Zhou, J.~Zhou, and H.~Yang, ``{{OFA}}: {{Unifying Architectures}}, {{Tasks}}, and {{Modalities Through}} a {{Simple Sequence-to-Sequence Learning Framework}},'' Jun. 2022.

\bibitem[Yu et~al.(2022)Yu, Wang, Vasudevan, Yeung, Seyedhosseini, and Wu]{yuCoCaContrastiveCaptioners2022}
J.~Yu, Z.~Wang, V.~Vasudevan, L.~Yeung, M.~Seyedhosseini, and Y.~Wu, ``{{CoCa}}: {{Contrastive Captioners}} are {{Image-Text Foundation Models}},'' https://arxiv.org/abs/2205.01917v2, May 2022.

\bibitem[Beliy et~al.(2019)Beliy, Gaziv, Hoogi, Strappini, Golan, and Irani]{beliyVoxelsPixelsBack2019}
R.~Beliy, G.~Gaziv, A.~Hoogi, F.~Strappini, T.~Golan, and M.~Irani, ``From voxels to pixels and back: {{Self-supervision}} in natural-image reconstruction from {{fMRI}},'' Jul. 2019.

\bibitem[Mozafari et~al.(2020)Mozafari, Reddy, and VanRullen]{mozafariReconstructingNaturalScenes2020}
M.~Mozafari, L.~Reddy, and R.~VanRullen, ``Reconstructing {{Natural Scenes}} from {{fMRI Patterns}} using {{BigBiGAN}},'' in \emph{2020 {{International Joint Conference}} on {{Neural Networks}} ({{IJCNN}})}, Jul. 2020, pp. 1--8.

\bibitem[Chen et~al.(2023)Chen, Qing, Xiang, Yue, and Zhou]{chenSeeingBrainConditional2023}
Z.~Chen, J.~Qing, T.~Xiang, W.~L. Yue, and J.~H. Zhou, ``Seeing {{Beyond}} the {{Brain}}: {{Conditional Diffusion Model With Sparse Masked Modeling}} for {{Vision Decoding}},'' in \emph{Proceedings of the {{IEEE}}/{{CVF Conference}} on {{Computer Vision}} and {{Pattern Recognition}}}, 2023, pp. 22\,710--22\,720.

\bibitem[Sun et~al.(2023)Sun, Li, Chen, Zhang, Wang, and Moens]{sunContrastAttendDiffuse2023}
J.~Sun, M.~Li, Z.~Chen, Y.~Zhang, S.~Wang, and M.-F. Moens, ``Contrast, {{Attend}} and {{Diffuse}} to {{Decode High-Resolution Images}} from {{Brain Activities}},'' https://arxiv.org/abs/2305.17214v2, May 2023.

\bibitem[Scotti et~al.(2024)Scotti, Tripathy, Villanueva, Kneeland, Chen, Narang, Santhirasegaran, Xu, Naselaris, Norman, and Abraham]{scottiMindEye2SharedSubjectModels2024}
P.~S. Scotti, M.~Tripathy, C.~K.~T. Villanueva, R.~Kneeland, T.~Chen, A.~Narang, C.~Santhirasegaran, J.~Xu, T.~Naselaris, K.~A. Norman, and T.~M. Abraham, ``{{MindEye2}}: {{Shared-Subject Models Enable fMRI-To-Image With}} 1 {{Hour}} of {{Data}},'' Mar. 2024.

\bibitem[Dosovitskiy et~al.(2021)Dosovitskiy, Beyer, Kolesnikov, Weissenborn, Zhai, Unterthiner, Dehghani, Minderer, Heigold, Gelly, Uszkoreit, and Houlsby]{dosovitskiy2021an}
\BIBentryALTinterwordspacing
A.~Dosovitskiy, L.~Beyer, A.~Kolesnikov, D.~Weissenborn, X.~Zhai, T.~Unterthiner, M.~Dehghani, M.~Minderer, G.~Heigold, S.~Gelly, J.~Uszkoreit, and N.~Houlsby, ``An image is worth 16x16 words: Transformers for image recognition at scale,'' in \emph{International Conference on Learning Representations}, 2021. [Online]. Available: \url{https://openreview.net/forum?id=YicbFdNTTy}
\BIBentrySTDinterwordspacing

\bibitem[Horikawa and Kamitani(2017)]{horikawaGenericDecodingSeen2017}
T.~Horikawa and Y.~Kamitani, ``Generic decoding of seen and imagined objects using hierarchical visual features,'' \emph{Nature Communications}, vol.~8, no.~1, p. 15037, May 2017.

\bibitem[Allen et~al.(2022)Allen, {St-Yves}, Wu, Breedlove, Prince, Dowdle, Nau, Caron, Pestilli, Charest, Hutchinson, Naselaris, and Kay]{allenMassive7TFMRI2022}
E.~J. Allen, G.~{St-Yves}, Y.~Wu, J.~L. Breedlove, J.~S. Prince, L.~T. Dowdle, M.~Nau, B.~Caron, F.~Pestilli, I.~Charest, J.~B. Hutchinson, T.~Naselaris, and K.~Kay, ``A massive {{7T fMRI}} dataset to bridge cognitive neuroscience and artificial intelligence,'' \emph{Nature Neuroscience}, vol.~25, no.~1, pp. 116--126, Jan. 2022.

\bibitem[Lin et~al.(2014)Lin, Maire, Belongie, Hays, Perona, Ramanan, Doll{\'a}r, and Zitnick]{lin2014microsoft}
T.-Y. Lin, M.~Maire, S.~Belongie, J.~Hays, P.~Perona, D.~Ramanan, P.~Doll{\'a}r, and C.~L. Zitnick, ``Microsoft coco: Common objects in context,'' in \emph{Computer Vision--ECCV 2014: 13th European Conference, Zurich, Switzerland, September 6-12, 2014, Proceedings, Part V 13}.\hskip 1em plus 0.5em minus 0.4em\relax Springer, 2014, pp. 740--755.

\bibitem[Goyal et~al.(2017)Goyal, Khot, Summers{-}Stay, Batra, and Parikh]{balanced_vqa_v2}
Y.~Goyal, T.~Khot, D.~Summers{-}Stay, D.~Batra, and D.~Parikh, ``Making the {V} in {VQA} matter: Elevating the role of image understanding in {V}isual {Q}uestion {A}nswering,'' in \emph{Conference on Computer Vision and Pattern Recognition (CVPR)}, 2017.

\bibitem[Ilharco et~al.(2021)Ilharco, Wortsman, Wightman, Gordon, Carlini, Taori, Dave, Shankar, Namkoong, Miller, Hajishirzi, Farhadi, and Schmidt]{ilharco_gabriel_2021_5143773}
\BIBentryALTinterwordspacing
G.~Ilharco, M.~Wortsman, R.~Wightman, C.~Gordon, N.~Carlini, R.~Taori, A.~Dave, V.~Shankar, H.~Namkoong, J.~Miller, H.~Hajishirzi, A.~Farhadi, and L.~Schmidt, ``Openclip,'' Jul. 2021, if you use this software, please cite it as below. [Online]. Available: \url{https://doi.org/10.5281/zenodo.5143773}
\BIBentrySTDinterwordspacing

\end{thebibliography}

\end{document}